\pdfoutput=1

\documentclass[11pt]{article}

\usepackage[]{emnlp2021}

\usepackage{times}
\usepackage{latexsym}
\usepackage{graphicx}
\usepackage{adjustbox}
\usepackage{booktabs}
\usepackage{todonotes}
\usepackage[ruled,vlined]{algorithm2e}
\usepackage[T1]{fontenc}

\usepackage[utf8]{inputenc}

\usepackage{microtype}
\usepackage{amssymb}
\usepackage{pifont}
\newcommand{\cmark}{\ding{51}}%
\title{Improving Unsupervised Question Answering via Summarization-Informed Question Generation}

 \author{
  \textbf{Chenyang Lyu}$^\dag$
  ~~~~ \textbf{Lifeng Shang}$^\ddag$~~~~ \textbf{Yvette Graham}$^\P$~~~~ \textbf{Jennifer Foster}$^\dag$~~~~  \\ \textbf{Xin Jiang}$^\ddag$~~~~ \textbf{Qun Liu}$^\ddag$ \\
  $^\dag$ School of Computing, Dublin City University, Dublin, Ireland \\
  $^\ddag$ Huawei Noah's Ark Lab,  Hong Kong, China \\
  $^\P$ School of Computer Science and Statistics, Trinity College Dublin, Dublin, Ireland\\
  \texttt{chenyang.lyu2@mail.dcu.ie}, \texttt{ygraham@tcd.ie}, \texttt{jennifer.foster@dcu.ie} \\
  \texttt{\{{shang.lifeng, jiang.xin, qun.liu\}@huawei.com}}
}

\begin{document}
\maketitle
\begin{abstract}

Question Generation (QG) is the task of generating a plausible question for a given \textit{<passage, answer>} pair.  \textit{Template-based QG}  uses linguistically-informed heuristics to transform declarative sentences into interrogatives, whereas \textit{supervised QG} uses existing Question Answering (QA) datasets to train a system to generate a question given a passage and an answer.
A disadvantage of the heuristic approach is that  the generated questions are heavily tied to their declarative counterparts. A disadvantage of the supervised approach is that they are heavily tied to the domain/language of the QA dataset used as training data.
In order to overcome these shortcomings, we propose an unsupervised QG method which uses questions generated heuristically from \textit{summaries} as a source of training data for a  QG system.  We make use of freely available news summary data, transforming declarative summary sentences into appropriate questions using heuristics informed by dependency parsing, named entity recognition and semantic role labeling. The resulting questions are then combined with the original news articles to train an end-to-end neural QG model. 
We extrinsically evaluate our approach using unsupervised QA: our QG model is used to  generate synthetic QA pairs for training a QA model. Experimental results show that, trained with only 20k English Wikipedia-based synthetic QA pairs, the QA model substantially outperforms previous unsupervised models on three in-domain datasets (SQuAD1.1, Natural Questions, TriviaQA) and three out-of-domain datasets (NewsQA, BioASQ, DuoRC), demonstrating the transferability of the approach. 
\end{abstract}

\section{Introduction}

The aim of Question Generation (QG) is the production of meaningful questions given a set of input passages and corresponding answers, a task with many applications including dialogue systems as well as education~\cite{education-2005}. 
Additionally, QG can be applied to Question Answering (QA) for the purpose of data augmentation~\cite{puri-etal-2020-training-synthetic} where labeled \textit{<passage, answer, question>} triples are combined with synthetic \textit{<passage, answer, question>} triples produced by a QG system to train a QA system, and unsupervised QA~\cite{lewis-etal-2019-unsupervised-cloze}, in which only the QG system output is used to train the QA system. 
\begin{figure}[t]
    \centering
    \includegraphics[scale=0.25]{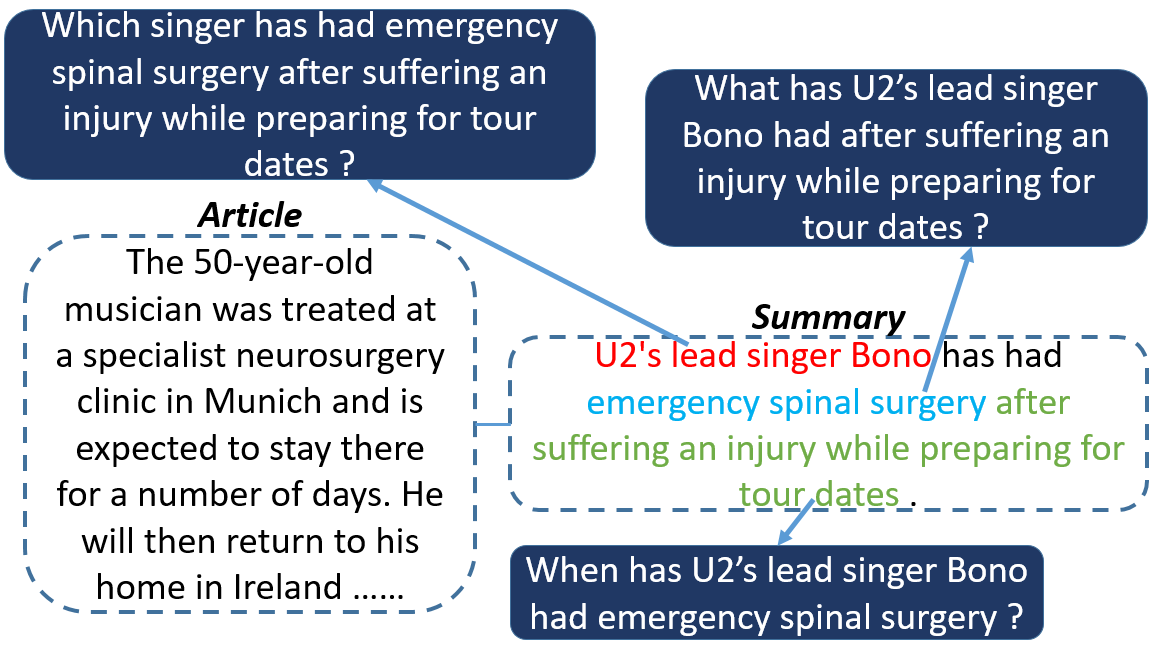}
    \caption{Example questions generated via heuristics informed by semantic role labeling of summary sentences using different candidate answer spans}
    \label{fig:introduction_graph}
\end{figure}


Early work on QG focused on template or rule-based approaches, employing syntactic knowledge to manipulate constituents in declarative sentences to form  interrogatives~\cite{Heilman2009Question-noah,heilman-smith-2010-good}. Although template-based methods are capable of generating linguistically correct questions, the resulting questions often lack variety and incur high lexical overlap with corresponding declarative sentences. 
For example, the question generated from the sentence \textit{\underline{Stephen Hawking} announced the party in the morning}, with \textit{Stephen Hawking} as the candidate answer span, could be \textit{Who announced the party in the morning?}, with a high level of lexical overlap between the generated question and the declarative sentence. This is undesirable in a QA system \cite{kang-etal-2020-handling} since the strong lexical clues in the question would make it a poor test of real comprehension. 

Neural seq2seq models~\cite{NIPS2014_seq2seq} have come to  dominate QG~\cite{du-etal-2017-learning}, and are commonly trained with \textit{<passage, answer, question>} triples taken from human-created QA datasets~\cite{dzendzik2021english_qa-dataset} and this limits applications to the  domain and language of datasets.
Furthermore, the process of constructing such datasets involves a significant investment of time and resources.


We subsequently propose a new unsupervised approach
that frames QG as a summarization-questioning process.

%
By employing freely available summary data, 
we firstly apply dependency parsing, named entity recognition and semantic role labeling to summaries, before applying a set of heuristics that generate questions based on parsed summaries. An end-to-end neural generation system is then trained employing the \textit{original news articles} as input and the heuristically generated questions as target output.

An example is shown in Figure~\ref{fig:introduction_graph}. The summary is used as a bridge between the questions and passages. Because the questions are generated from the summaries and not from the original passages, they have less of a lexical overlap with the passages. Crucially, however, they remain semantically close to the passages since the summaries by definition contain the most important information contained in the passages. A second advantage of this QG approach is that it does not rely on the existence of a QA dataset, and it is arguably easier to obtain summary data in a given language than equivalent QA data since summary data is created for many purposes (e.g. news, review and thesis summaries) whereas many QA datasets are created specifically for training a QA system.


In order to explore the effectiveness of our method, we carry out extensive experiments. We provide an extrinsic evaluation, and train an English QG model using news summary data. We employ our QG model to generate synthetic QA data to train a QA model in an unsupervised setting and test the approach with six English QA datasets: SQuAD1.1, Natural Questions, TriviaQA, NewsQA, BioASQ and DuoRC~\cite{rajpurkar-etal-2016-squad1.1, kwiatkowski-etal-2019-naturalquestions-nq, joshi-etal-2017-triviaqa, trischler-etal-2017-newsqa, bio_asq, saha-etal-2018-duorc}. Experiment results show that our approach substantially improves over previous unsupervised QA models even when trained on substantially fewer synthetic QA examples.

Our contributions can be summarized as follows:
\begin{enumerate}
\setlength\itemsep{-0.5em}
\item We propose a novel unsupervised QG approach that employs summary data and syntactic/semantic analysis, which to our best knowledge is the first work connecting text summarization and question generation in this way;

\item We employ our QG model to generate synthetic QA data achieving state-of-the-art performance even at low volumes of synthetic training data.
\end{enumerate}
\section{Related Work}
\begin{figure*}
    \centering
    \includegraphics[scale=0.2705]{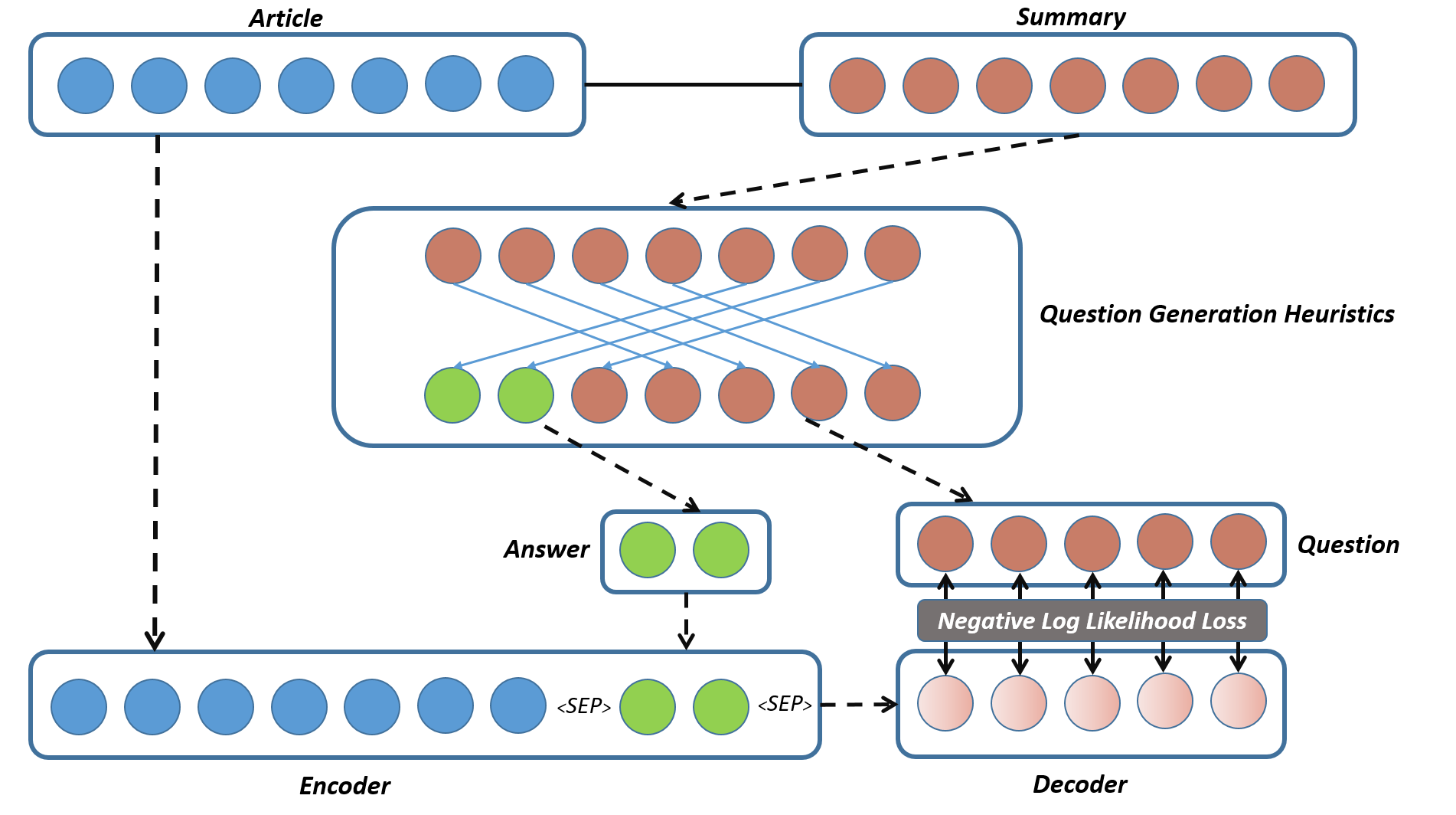}
    \caption{An overview of our approach where \textit{Answer} and \textit{Question} are generated based on \textit{Summary} by the \textit{Question Generation Heuristics}, the \textit{Answer} is combined with the \textit{Article} to form the input to the Encoder, the \textit{Question} is employed as the ground-truth label for the outputs of the Decoder.}
    \label{fig_main_approach}
\end{figure*}
\paragraph{Question Generation} 
Traditional approaches to QG mostly employ linguistic templates and rules to transform declarative sentences into interrogatives \cite{Heilman2009Question-noah}. Recently, \newcite{dhole-manning-2020-syn-qg} showed that, with the help of advanced neural syntactic parsers, template-based methods are capable of generating high-quality questions from texts. 

Neural seq2seq generation models have additionally been widely employed in QG, with QG data usually borrowed from existing QA datasets~\cite{du-etal-2017-learning,sun-etal-2018-answer-yanjunma,ma2020improving}. Furthermore, reinforcement learning has been employed by \newcite{zhang-bansal-2019-addressing,chen2020reinforcement,xie-etal-2020-exploring} to directly optimize discrete evaluation metrics such as BLEU \cite{papineni-etal-2002-bleu}. \newcite{lewis-etal-2020-bart} and \newcite{song2019mass} show that a large-scale pre-trained model can achieve state-of-the-art performance for supervised QG \cite{dong2019unified-unilm,narayan2020qurious-maji}.


\paragraph{Question Generation Evaluation} BLEU~\cite{papineni-etal-2002-bleu}, ROUGE~\cite{lin-2004-rouge} and Meteor~\cite{banerjee-lavie-2005-meteor} metrics are commonly borrowed from text generation tasks to evaluate QG. Even with respect to original text generation tasks, however, the use of such metrics has been questioned~\cite{callison-burch-etal-2006-evaluating,10.1162/coli_a_00322}. Such metrics are particularly problematic for QG evaluation since multiple plausible questions exist for a given passage and answer. Consequently, there has been a shift in focus to evaluating QG using an extrinsic evaluation  that generates synthetic QA pairs for the purpose of evaluating their effectiveness as a data augmentation or unsupervised QA approach~\cite{alberti-etal-2019-synthetic, puri-etal-2020-training-synthetic,shakeri-etal-2020-end-wangzhiguo}.



\paragraph{Unsupervised QA} 
In unsupervised QA, the QA model is trained using synthetic data based on a QG model instead of an existing QA dataset. Instead of resorting to existing QA datasets, unsupervised QG methods have been employed, such as Unsupervised Neural Machine Translation~\cite{lewis-etal-2019-unsupervised-cloze}. \newcite{fabbri-etal-2020-template} and \newcite{li-etal-2020-harvesting} propose template/rule-based methods for generating questions and employ retrieved paragraphs and cited passages as source passages to alleviate the problems of lexical similarities between passages and questions. \newcite{alberti-etal-2019-synthetic,puri-etal-2020-training-synthetic,shakeri-etal-2020-end-wangzhiguo} additionally employ existing QA datasets to train a QG model. Although related, this work falls outside the scope of unsupervised QA. 

\section{Methodology}
 Diverging from supervised neural question generation models trained on existing QA datasets, the approach we propose employs synthetic QG data, that we create from summary data using a number of heuristics, to train a QG model. We provide an overview of the proposed method is shown in Figure~\ref{fig_main_approach}.
We then employ the trained QG model to generate synthetic QA data that is further employed to train an unsupervised QA model.

\subsection{Question Generation}
\label{3.1-parsing-summary}
In order to avoid generating trivial questions that are highly similar to corresponding declarative statements, we employ summary data as a bridge connecting the generated question and the original article.\footnote{Data we employ in experiments is news summary data originally from BBC News \cite{narayan-etal-2018-xsum} and the news articles are typically a few hundred words in length.}
The process we employ involves, firstly Dependency Parsing (DP) of summary sentences, followed by Named-Entity Recognition (NER) and finally Semantic Role Labeling (SRL). DP is firstly employed as a means of  identifying the main verb (root verb), in addition to other constituents such as auxiliaries. NER is then responsible for tagging all entities in the summary sentence to facilitate discovery of the most appropriate question words to generate. The pivotal component of linguistic analysis is then SRL, employed to obtain all semantic frames for the summary sentence. Each frame consists of a verb followed by a set of arguments which correspond to phrases in the sentence. An argument could comprise, for example, an \textit{Agent} (who initiates the action described by the verb), a \textit{Patient} (who undergoes the action), and a set of modifier arguments such as a temporal \textit{ARG-TMP} or locative argument \textit{ARG-LOC}. 
Questions are then generated from the arguments according to argument type and NER tags, which means that wh-words can be determined jointly.

Returning to the example in Figure~\ref{fig:introduction_graph}: 
given the SRL analysis \textit{ [U2's lead singer Bono \textbf{ARG-0}] has [had \textbf{VERB}] [emergency spinal surgery \textbf{ARG-1}] [after suffering an injury while preparing for tour dates \textbf{ARG-TMP}].}, the three questions shown in Figure~\ref{fig:introduction_graph} can be generated based on these three arguments.

The pseudocode for our algorithm to generate questions is shown in Algorithm~\ref{qg_xsum}.
\begin{algorithm}[!]
\small
\SetAlgoLined

$ S = summary $

$ srl\_frames = SRL(S)$

$ ners = NER(S)$

$ dps = DP(S)$

$examples = []$

 \For{frame in srl\_frames}{
  
  $root\_verb = dps_{root}$
  
  $verb = frame_{verb}$
  
  \If{root\_verb \textbf{equal to} verb}{

      \For{arg in frame}{
        $wh* = identify\_wh\_word(arg, ners)$ \\
        
        $base\_verb, auxs = decomp\_verb(arg, dps, root\_verb)$ \\
        
        $Q_{arg} = wh\_move(S, wh*, base\_verb, auxs)$
        
        $Q_{arg} = post\_edit(Q_{arg})$
        
        $examples.append(context, Q_{arg}, arg)$
        
      }
   }
 }
 \caption{Question Generation Heuristics}
 \label{qg_xsum}
\end{algorithm}
We first obtain all dependency edges and labels~($dps$), NER tags~($ners$) and SRL frames~($srl\_frames$) of a summary sentence. 
We then iterate through all arguments in the frame of the $root\_verb$ (the verb whose dependency label is $root$) and identify appropriate wh-words~($wh*$) for each argument using the function $identify\_wh\_word$ 
according to its argument type and the NER tags of entities in the argument. We follow \newcite{dhole-manning-2020-syn-qg} to use the standard wh-words in English associated with appropriate argument types and NER tags. We then decompose the current main verb to its base form~($base\_verb$) and appropriate auxiliary words~($auxs$) in the $decomp\_verb$ function, before finally inserting the wh-words and the auxiliary verbs in appropriate positions  using the $wh\_move$. 
As can be seen from Figure~\ref{qg_xsum}, a single summary sentence generates multiple questions when its SRL frame has multiple arguments.

\subsection{Training a Question Generation Model}
\label{3.2-train-question-generation}

The summarization data we employ consists of \textit{<passage-summary>} pairs.
Questions are generated from the summaries using the  heuristics described in Section~\ref{3.1-parsing-summary}, so that we have \textit{<passage-summary>} pairs and \textit{<summary-question-answer>} triples, which we then combine to form \textit{<passage-answer-question>} triples to train a QG model. We train an end-to-end seq2seq model rather than deploying a pipeline in which the summary is first generated followed by the question to eliminate the risk of error accumulation in the generation process. By using this QG data to train a neural generation model, we expect the model to learn a combination of summarization and question generation. In other words, such knowledge can be implicitly injected into the neural generation model via our QG data.

To train the question generation model, we concatenate each passage and answer to form a sequence: \textit{passage <SEP> answer <SEP>}, where \textit{<SEP>} is a special token used to separate the passage and answer. This sequence is the input and the question is the target output (objective).
In our experiments, we use BART~\cite{lewis-etal-2020-bart} for generation, which is optimized by the following negative log likelihood loss function:
\begin{equation}
    L = -\sum_{i=1}^{N}log P(q_i|C, A)
\end{equation}
where $q_i$ is the \textit{i}-th token in the question, and $C$ and $A$ are context and answer, respectively. 

\section{Experiments}
We test our idea of using summaries in question generation by 
applying the questions generated by our QG system in unsupervised QA. We describe the details of our experiment setup, followed by our unsupervised QA results on six English benchmark extractive QA datasets.

\subsection{Experiment Setup}
\subsubsection{Question Generation}
\label{question_generation_experiments}
\paragraph{Datasets}
We test the proposed method using news summary data from XSUM~\cite{narayan-etal-2018-xsum}, crawled from BBC news website.\footnote{\url{www.bbc.com}} XSUM contains 226,711 \textit{<passage-summary>} pairs, with each summary containing a single sentence.
\paragraph{QG  Details}
We employ AllenNLP\footnote{https://demo.allennlp.org/}~\cite{Gardner2017AllenNLP} to obtain dependency trees, named entities and semantic role labels for  summary sentences, before further employing this knowledge to generate questions from summaries  following the algorithm described in Section~\ref{3.1-parsing-summary}. We remove any generated \textit{<passage-answer-question>} triples that meet one or more of the following three conditions: 
\begin{enumerate}
\setlength\itemsep{-0.5em}
\item Articles longer than 480 tokens (exceeding the maximum BART input length);
\item Articles in which fewer than 55\% of tokens in the answer span are not additionally present in the passage (to ensure sufficient lexical overlap between the answer and passage);
\item  Questions shorter than 5 tokens (very short questions are likely to have removed too much information)
\end{enumerate}
For the dataset in question, this process resulted in a total of 14,830 \textit{<passage-answer-question>} triples.

For training the QG model, we employ implementations of BART~\cite{lewis-etal-2020-bart} from Huggingface~\cite{Wolf2019HuggingFacesTS}. The QG model we employ is BART-base. We train the QG model on the QG data for 3 epochs with a learning rate of $3 \times 10^{-5}$, using the  AdamW optimizer~\cite{adamw}.

\subsubsection{Unsupervised QA}\label{qa-exp-setup}
\label{synthetic_qa_generation}
\paragraph{Datasets}
We carry out experiments on six extractive QA datasets, namely, SQuAD1.1~\cite{rajpurkar-etal-2016-squad1.1}, NewsQA~\cite{trischler-etal-2017-newsqa}, Natural Questions~\cite{kwiatkowski-etal-2019-naturalquestions-nq}, TriviaQA~\cite{joshi-etal-2017-triviaqa}, BioASQ~\cite{bio_asq} and DuoRC~\cite{saha-etal-2018-duorc}. We employ the official data of SQuAD1.1, NewsQA and TriviaQA and for Natural Questions, BioASQ and DuoRC, we employ the pre-processed data released by MRQA~\cite{fisch2019mrqa}.

\paragraph{Unsupervised QA Training Details}
To generate synthetic QA training data, we make use of Wikidumps \footnote{https://dumps.wikimedia.org/} by firstly removing all HTML tags and reference links, then  extracting paragraphs that are longer than 500 characters, resulting in 60k paragraphs sampled from all paragraphs of Wikidumps. We employ the NER toolkits of Spacy\footnote{https://spacy.io/}~\cite{spacy} and AllenNLP\footnote{https://demo.allennlp.org/named-entity-recognition/named-entity-recognition}~\cite{Gardner2017AllenNLP} to extract entity mentions in the paragraphs. We then remove \textit{paragraph, answer} pairs that meet one or more of the following three conditions:  1) paragraphs with less than 20 words and more than 480 words; 2) paragraphs with no extracted answer, or where the extracted answer is not in the paragraph due to text tokenization; 3) answers consisting of a single pronoun.

Paragraphs and answers are concatenated to form sequences of the form \textit{passage <SEP> answer <SEP>}, before being fed into the trained BART-QG model to obtain  corresponding questions. This results in 20k synthetic QA pairs, which are then employed to train an  unsupervised QA model. 

The QA model we employ is BERT-large-whole-word-masking (which we henceforth refer to as BERT-large for ease of reference). Document length and stride length are 364 and 128 respectively, the learning rate is set to $1 \times 10^{-5}$. Evaluation metrics for unsupervised QA are Exact Match~(EM) and F-1 score.

\begin{table}
  \centering
  \scalebox{1}{
  \begin{tabular}{lcccc}
    \toprule
     && \multicolumn{2}{c}{SQuAD1.1}  \\ [1ex]
    
     Models  && EM &  F-1   \\
     
    \midrule
    
    \textsc{Supervised Models} \\
    
    \hspace{0.2cm} Match-LSTM  && 64.1  & 73.9   \\
    \hspace{0.2cm} BiDAF  && 66.7 & 77.3   \\
    
    \hspace{0.2cm} BERT-base  && 81.2 & 88.5 \\
    
    \hspace{0.2cm} BERT-large  && 84.2 & 91.1  \\
    
    \hline
    
    \textsc{Unsupervised Models} \\
    
    \hspace{0.2cm} \newcite{lewis-etal-2019-unsupervised-cloze}  && 44.2 & 54.7  \\
    
    \hspace{0.2cm} \newcite{li-etal-2020-harvesting}  && 62.5 & 72.6   \\

    \hspace{0.2cm} Our Method && \textbf{65.6} & \textbf{74.5}  \\ 
    \bottomrule
    \end{tabular}%
    }
    \caption{In-domain experimental results of supervised and unsupervised methods on SQuAD1.1. 
    The highest  scores of unsupervised methods are in bold.
    }
  \label{uqa_results_1}%
\end{table}

\subsection{Results}


We use the 20k generated synthetic QA pairs to train a BERT QA model and first validate its performance on the development sets of three benchmark QA datasets based on Wikipedia -- SQuAD1.1, Natural Questions and TriviaQA. The results of our method are shown in Tables~\ref{uqa_results_1} and \ref{uqa_results_2}. 
 The unsupervised baselines we compare with are as follows:
 \begin{enumerate}
 \setlength\itemsep{-0.5em}
 \item \newcite{lewis-etal-2019-unsupervised-cloze} employ unsupervised neural machine translation~\cite{artetxe2018unsupervised-unmt} to train a QG model; 4M synthetic QA examples were generated to train a QA model;
 \item \newcite{li-etal-2020-harvesting} employ dependency trees to generate questions and employed cited documents as passages.
 \end{enumerate}
For comparison, we also show the results of some supervised models fine-tuned on the corresponding training sets: Match-LSTM~\cite{wang2016machine_match-lstm}, BiDAF~\cite{seo2016bidirectional_bidaf}, BERT-base and BERT-large~\cite{bert}.

SQuAD1.1 results are shown in  Table~\ref{uqa_results_1}. The results of all baseline models are taken directly from published work.
As can be seen from results in Table~\ref{uqa_results_1}, our proposed method outperforms all unsupervised baselines, and even exceeds the performance of one supervised model, Match-LSTM~\cite{wang2016machine_match-lstm}.

Results for Natural Questions and TriviaQA are shown in Table~\ref{uqa_results_2}. The results of all baseline models were produced 

using the released synthetic QA data to finetune a BERT-large model.
 Our method outperforms previous state-of-the-art unsupervised methods by a substantial margin, obtaining relative improvements over the best unsupervised baseline model of 47\% with respect to EM, 10\% F-1 on Natural Questions, and by 34\% EM and 12\% F-1 on TriviaQA.

In summary, our method achieves the best performance (both in terms of EM and F-1) out of three unsupervised models on all three tested datasets. Furthermore, this high performance is possible with as few as 20k training examples. Compared to previous work, this is approximately less than 10\% of the training data employed~\cite{li-etal-2020-harvesting}.

\begin{table}
\setlength{\tabcolsep}{2.5pt}
  \centering
  \scalebox{0.9}{
  \begin{tabular}{lcccccc}
    \toprule
    && \multicolumn{2}{c}{NQ}  && \multicolumn{2}{c}{TriviaQA} \\ [1ex]
    
     Models  && EM &  F-1 &&  EM &  F-1  \\
     
    \midrule
    
    \textsc{Supervised Models} \\

    \hspace{0.2cm} BERT-base  && 66.1  & 78.5 && 65.1  & 71.2  \\
    
    \hspace{0.2cm} BERT-large  && 69.7  & 81.3 && 67.9  & 74.8  \\
    
    \hline
    
    \textsc{Unsupervised Models} \\
    
    \hspace{0.2cm} \newcite{lewis-etal-2019-unsupervised-cloze}  && 27.5  & 35.1 && 19.1  & 23.8  \\
    
    \hspace{0.2cm} \newcite{li-etal-2020-harvesting} && 31.3  & 48.8 && 27.4  & 38.4  \\
    
    
    \hspace{0.2cm} Our Method && \textbf{46.0}  & \textbf{53.5} && \textbf{36.7}  & \textbf{43.0} \\ 
    \bottomrule
    \end{tabular}%
    }
    \caption{In-domain experimental results: Natural Questions and TriviaQA.}
  \label{uqa_results_2}%
\end{table}
\begin{table}
\footnotesize
\setlength{\tabcolsep}{2pt}
  \centering
  \scalebox{1.05}{
  \begin{tabular}{lcccccccccc}
    \toprule
    &&  \multicolumn{2}{c}{NewsQA} && \multicolumn{2}{c}{BioASQ}  && \multicolumn{2}{c}{DuoRC} \\ [1ex]
    
     && EM &  F-1  &&  EM &  F-1 &&  EM &  F-1  \\
     
    \midrule
    
    
    \newcite{lewis-etal-2019-unsupervised-cloze} && 19.6 & 28.5 && 18.9  & 27.0 && 26.0  & 32.6  \\
    
    \newcite{li-etal-2020-harvesting} && 33.6 & 46.3 && 30.3  & 38.7 && 32.7  & 41.1  \\

    
    Our Method && \textbf{37.5} & \textbf{50.1} && \textbf{32.0}  & \textbf{43.2} && \textbf{38.8}  & \textbf{46.5} \\ 
    
    \bottomrule
    \end{tabular}%
    }
    \caption{Out-of-domain experimental results of unsupervised methods on NewsQA, BioASQ and DuoRC. The results of two baseline models on NewsQA are taken from~\newcite{li-etal-2020-harvesting} and their results on BioASQ and DuoRC are from fine-tuning a BERT-large model on their synthetic data.
    }
  \label{uqa_results_out_domain}%
\end{table}

\paragraph{Transferability of Our Generated Synthetic QA Data}
We also validate our method's efficacy on three out-of-domain QA datasets: NewsQA created from news articles, BioASQ created from biomedical articles, and DuoRC created from movie plots, for the purpose of evaluating the transferability of the Wikipedia-based  synthetic data. Results in Table~\ref{uqa_results_out_domain} show that our proposed method additionally outperforms the  unsupervised baseline models on the out-of-domain datasets, achieving F1 improvements over previous state-of-the-art methods by 3.8, 4.5 and 5.4 points respectively. 
It is worth noting that our data adapts very well to DuoRC, created from movie plots where the narrative style is expected to require  more complex reasoning.
Experiment results additionally indicate that our generated synthetic data transfers 
well to domains distinct from  that of the original summary data. 

\section{Analysis}
\begin{table*}
\setlength{\tabcolsep}{0.5pt}
  \centering
  \scalebox{0.96}{
  \begin{tabular}{lllll}
    \toprule
    
    Questions && Answer && Comments \\
     
    \midrule
    
    \textit{who is the frontman of swedish rock band mhiam ?}  && \textit{Mattis Malinen} &&  \cmark\\
    
    \textit{which sultan  has been in bosnia for more than a year ?} && \textit{Sultan Mehmed II} &&  \cmark \\
    
    \textit{what is a major economic driver for the state of ohio ?} && \textit{Ohio's geographic location} && \cmark\\
    
    \textit{in what time was the first parish council elected ?} && \textit{March 1972} &&  \cmark \\
    
    \textit{what do the chattanooga area will host in 2017 ?} && \textit{the Ironman Triathlon} &&  \cmark grammar error  \\
    
    \textit{what have sold five cars in the uk this year ?} && \textit{Surrey Motors} &&  missing information\\
    
    \textit{when did the first military college in the us open  ?} && \textit{2009} && factual error  \\
    
    \textit{what has been described as a " giant fish " ?} && \textit{Darwin} && mismatch  \\
    
    \bottomrule
    \end{tabular}%
    
    }
    \caption{Examples of generated questions with corresponding answers.\cmark represents correct examples.}
  \label{uqg_examples}%
\end{table*}








\subsection{Effect of Answer Extraction}
\begin{table*}
\footnotesize
\setlength{\tabcolsep}{3pt}
  \centering
  \scalebox{1.2}{
  \begin{tabular}{lccccccccccccc}
    \toprule
    && \multicolumn{2}{c}{SQuAD1.1} && \multicolumn{2}{c}{NewsQA} && \multicolumn{2}{c}{NQ}  && \multicolumn{2}{c}{TriviaQA} \\ [1ex]
    
     Models && EM &  F-1  &&  EM &  F-1 &&  EM &  F-1  &&  EM &  F-1\\
     
    \midrule
    
    Our Method~(NER-extracted answers)\dag && 65.6 & 74.5 && 37.5 & 50.1 && 46.0  & 53.5 && 36.7  & 43.0  \\
    
    Our Method~(Human-extracted answers) \ddag && \textbf{68.0} & \textbf{79.5}  && \textbf{40.5} & \textbf{59.3} && \textbf{57.3} & \textbf{66.7} && \textbf{54.2}  & \textbf{61.1} \\
    
    \bottomrule
    \end{tabular}%
    }
    \caption{Comparison between synthetic data generated based on Wikipedia and synthetic data generated based on corresponding training set. \dag are results of QA model finetuned on synthetic data generated based on NER-extracted answers, \ddag are results of QA model finetuned on synthetic data based on the answers in the training set of SQuAD1.1, NewsQA, NQ and TriviaQA.
    }
  \label{uqa_results_comp_ans}%
\end{table*}

In the unsupervised QA experiments, we extracted answers from Wikipedia passages before feeding them into our QG model to obtain questions. These \textit{<passage, answer, question>} triples constitute the synthetic data employed to train the QA model. 
Additionally, we wish to consider what might happen if we instead employ passages and answers taken directly from the QA training data? 
Doing this would mean that the QA system is no longer considered unsupervised but  
we carry out this experiment
in order to provide insight into the degree to which there may be room for improvement in terms of our NER-based automatic answer extraction method (described in Section~\ref{qa-exp-setup}). 
For example, there could well be a gap between the NER-extracted answers and human-extracted answers, and in this case, the NER could extract answers, for example, that are not entirely worth asking about or indeed miss answers that are highly likely to be asked about. Results of the two additional settings are shown in Table~\ref{uqa_results_comp_ans} -- answer extraction has quite a large effect on the quality of generated synthetic QA data.  When we employ the answers from the training set, the performance of the QA model is improved by 
5 F-1 points for SQuAD1.1, and over 10  F-1 points for Natural Questions and TriviaQA.


\subsection{Effect of Different Heuristics}

We additionally investigate the effect of a range of alternate heuristics employed in the process of constructing the QG training data described in Section~\ref{3.1-parsing-summary}. Recall that the QG data is employed to train a question generator which is then employed to generate synthetic QA data for unsupervised QA. 

The heuristics are defined as follows:
\begin{itemize}
\setlength\itemsep{-0.5em}
\item \textbf{\texttt{Naive-QG}} only employs summary sentences as passages (instead of the original articles) and generates trivial questions in which only the answer spans are replaced with the appropriate question words. For example, for the sentence \textit{Stephen Hawking announced the party in the morning}, with \textit{the party} as the answer span, the question generated by \texttt{Naive-QG} would be \textit{Stephen Hawking announced what in the morning?} We employ the summary sentences as input and questions as target output to form the QG training data.
\item \textbf{\texttt{Summary-QG}} makes use of the original news articles of the summaries as passages rather than summary sentences to  avoid high lexical overlap between the passage and question.
\begin{table}
\footnotesize
  \centering
  \scalebox{1.2}{
  \begin{tabular}{lcccc}
    \toprule
    Heuristics && EM && F-1 \\ [1ex]
     
    \midrule
    
    Naive-QG  && 31.1 && 43.3  \\
    
    Summary-QG  && 50.9 && 59.4  \\
    \hspace{0.1cm} +Main Verb  && 53.8 && 63.6  \\
    \hspace{0.1cm} +Wh-Movement  && 59.5 && 67.7  \\
    \hspace{0.1cm} +Decomp-Verb  && 64.1 && 73.9  \\
    \hspace{0.1cm} +NER-Wh  && \textbf{65.4} && \textbf{74.8}  \\
    
    \bottomrule
    \end{tabular}%
    }
    \caption{Experiment results of the effects to unsupervised QA performance on SQuAD1.1 of using different heuristics in constructing QG data.
    }
  \label{uqa_results_comp_heuristics}%
\end{table}

\texttt{\textbf{Summary-QG}} can work with the following heuristics:
\begin{itemize}
\item \textbf{\texttt{Main Verb}}: we only generate questions based on the SRL frame of the main verb (root verb) in the dependency tree of the summary sentences, rather than using verbs in subordinate clauses; 
\item \textbf{\texttt{Wh-Movement}}: we move the question words to the beginning of the sentence. For example, in the sentence \textit{Stephen Hawking announced what in the morning?} we move \textit{what} to the beginning to obtain \textit{what Stephen Hawking announced in the morning?};
\item \textbf{\texttt{Decomp-Verb}}: the main verb is decomposed to its base form and auxiliaries;
\item \textbf{\texttt{NER-Wh}}: we employ the NER tags to get more precise question words for an answer. For example, for the answer span \textit{NBA player Michael Jordan}, the question words would be \textit{which NBA player}  instead of \textit{who} or \textit{what}.
\end{itemize}
\end{itemize}

We employ the QG data generated by these heuristics to train QG models, which leads to six BART-QG models. We then employ these six models to further generate synthetic QA data based on the same Wikipedia data and compare their performances on the SQuAD1.1 dev set.  
The results in Table~\ref{uqa_results_comp_heuristics} show that using articles as passages to avoid lexical overlap with their summary-generated questions greatly improves QA performance. \texttt{Summary-QG} outperforms \texttt{Naive-QG} by roughly 20 EM points  and 16 F-1 points. The results for the other heuristics show that they  continuously improve the performance, especially \texttt{Wh-Movement} and \texttt{Decomp-Verb} which make the questions in the QG data more similar to the questions in the QA dataset.

     
    
    
    

\begin{figure}[h]
    \centering
    \includegraphics[scale=0.52]{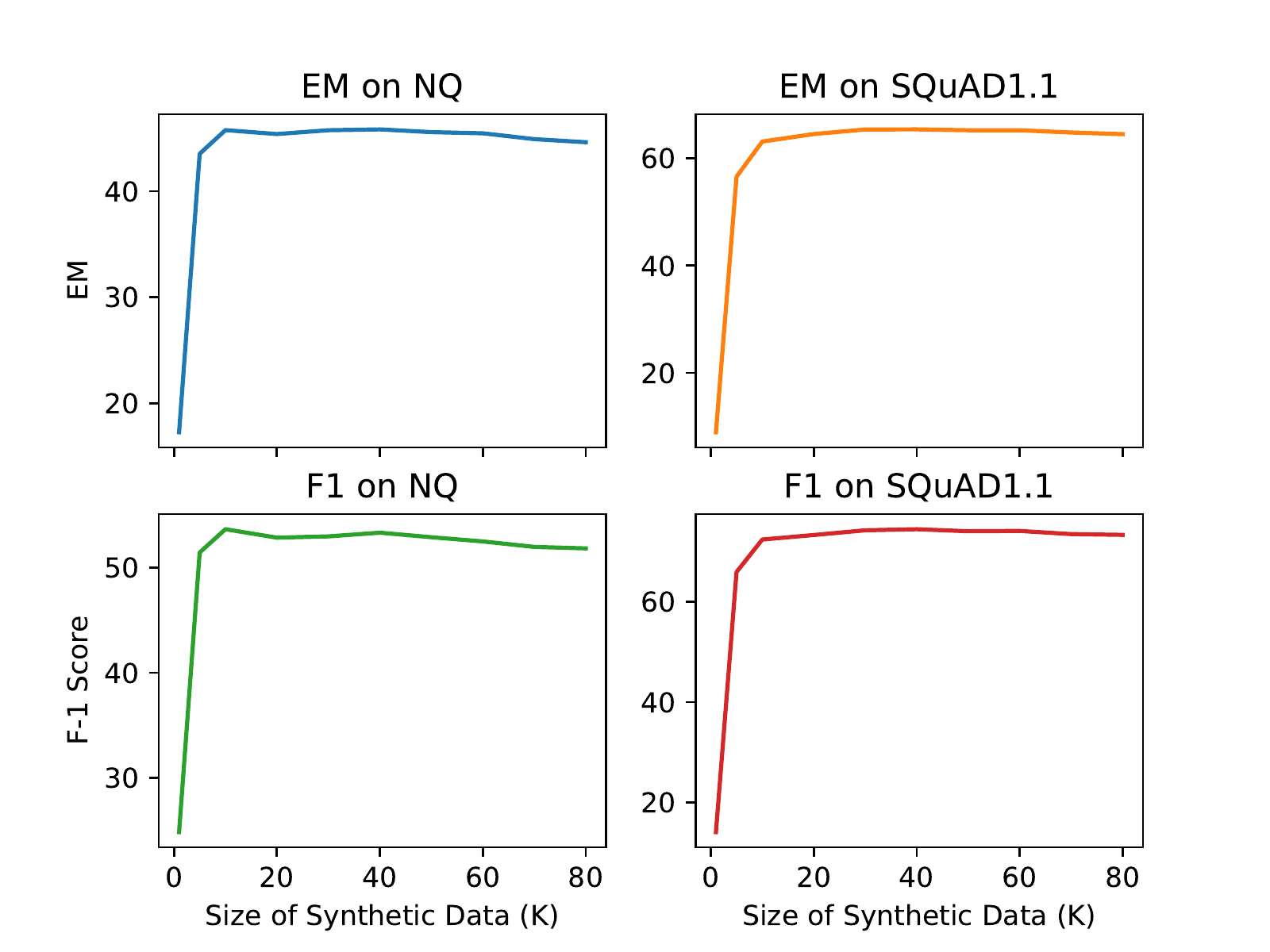}
    \caption{Experimental results on NQ and SQuAD1.1 of using different amount of synthetic data.}
    \label{fig:effect_of_data_size}
\end{figure}

\subsection{Effect of the Size of Synthetic QA Data}
We investigate the effects of varying the quantity of synthetic QA data.
Results in Figure~\ref{fig:effect_of_data_size} show that our synthetic data allows the QA model  to achieve competitive performance even with fewer than 20k examples, which suggests that our synthetic data contains sufficient QA knowledge to enable models to correctly answer a question with less synthetic data compared to previous unsupervised methods. 
The data-efficiency of our approach increases the feasibility of training a QA system for a target domain where there is no labeled QA data available.

\begin{figure}[h]
    \centering
    \includegraphics[scale=0.52]{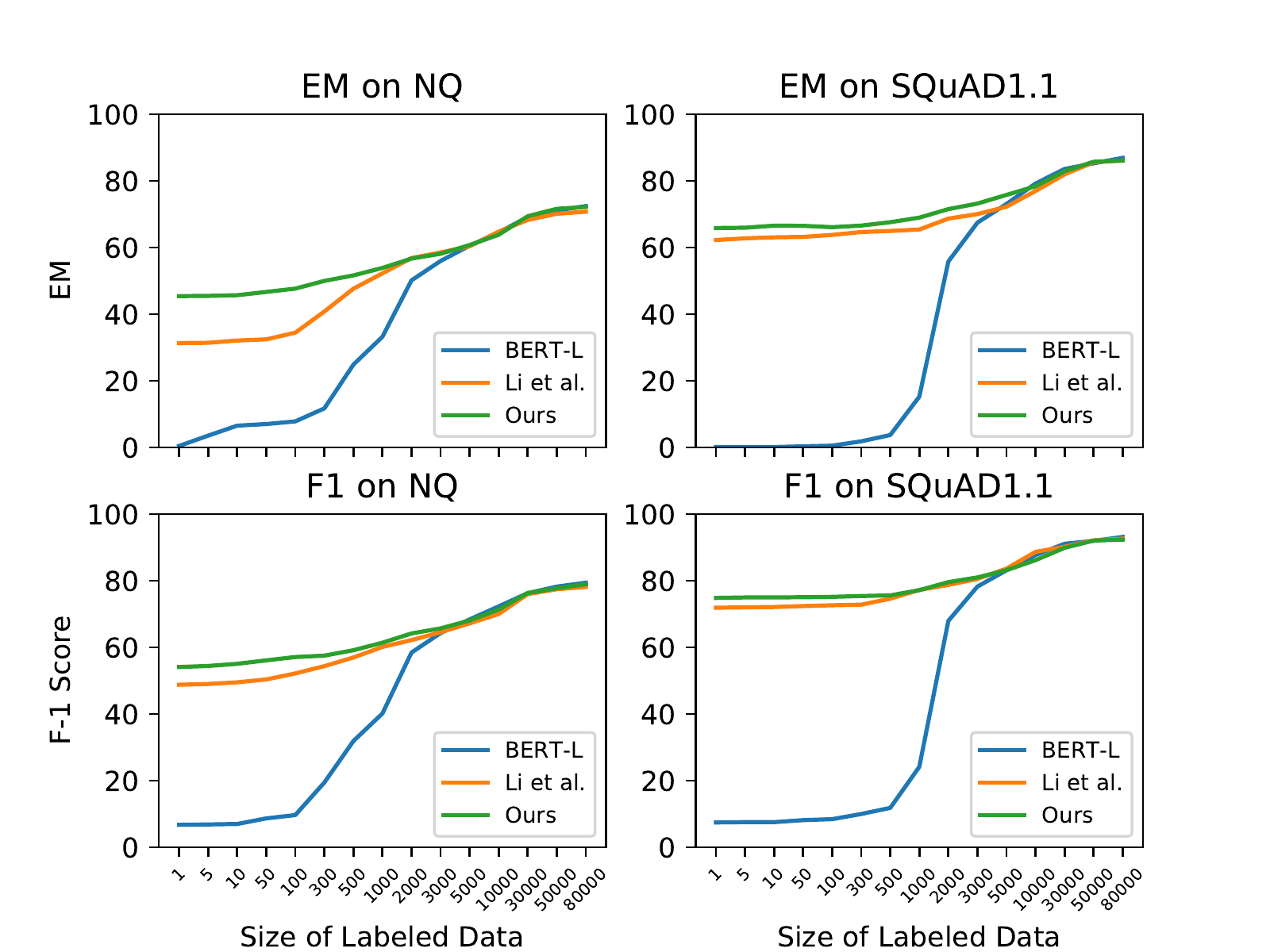}
    \caption{Experimental results of our method with comparison of \newcite{li-etal-2020-harvesting} and BERT-large using different amount of labeled QA examples in the training set of NQ and SQuAD1.1.}
    \label{fig:few_shot_learning}
\end{figure}

\subsection{Few-shot Learning}

We conduct experiments in a few-shot learning setting, in which we employ a limited number of labeled QA examples from the training set. We take the model trained with our synthetic QA data, the model trained with the synthetic QA data of \newcite{li-etal-2020-harvesting} and a vanilla BERT model, with all QA models employing BERT-large~\cite{bert}. We train these models using progressively increasing amounts of labeled QA samples from Natural Questions (NQ) and SQuAD1.1 and  assess their performance on corresponding dev sets. Results are shown in Figure~\ref{fig:few_shot_learning} where 
with only a small amount of labeled data~(less than 5,000 examples), our method outperforms \newcite{li-etal-2020-harvesting} and BERT-large, clearly demonstrating the efficacy of our approach in a few-shot learning setting.


\subsection{QG Error Analysis}
Despite substantial improvements over baselines, our proposed approach inevitably still incurs error and we therefore take a closer look at the questions generated by our QG model. We manually examine 50 randomly selected questions, 31 (62\%) of which were deemed high quality questions.  
The remaining 19 contain various errors with some  questions containing more than one error, including mismatched wh-word and answer (12\%), missing information needed to locate the answer (8\%), factual errors (10\%) and grammatical errors (8) (16\%)
Typical examples are shown in Table~\ref{uqg_examples}.

\section{Conclusion}
We propose an unsupervised question generation method which uses summarization data  to 1) minimize the lexical overlap between passage and question 
and 2) provide a QA-dataset-independent way of generating questions. Our unsupervised QA extrinsic evaluation on SQuAD1.1, NQ and TriviaQA using synthetic QA data generated by our method shows that our method substantially outperforms previous methods for generating synthetic QA for unsupervised QA. Furthermore, our synthetic QA data transfers well to the out-of-domain datasets. 
Future work includes refining our question generation heuristics and applying our approach to other languages.
\section*{Acknowledgements}

This work was funded by Science Foundation Ireland through the SFI Centre for Research Training in Machine Learning (18/CRT/6183). We also thank the  reviewers for their insightful and helpful comments.

\bibliography{anthology,custom}

\begin{thebibliography}{44}
\expandafter\ifx\csname natexlab\endcsname\relax\def\natexlab#1{#1}\fi

\bibitem[{Alberti et~al.(2019)Alberti, Andor, Pitler, Devlin, and
  Collins}]{alberti-etal-2019-synthetic}
Chris Alberti, Daniel Andor, Emily Pitler, Jacob Devlin, and Michael Collins.
  2019.
\newblock \href {https://doi.org/10.18653/v1/P19-1620} {Synthetic {QA} corpora
  generation with roundtrip consistency}.
\newblock In \emph{Proceedings of the 57th Annual Meeting of the Association
  for Computational Linguistics}, pages 6168--6173, Florence, Italy.
  Association for Computational Linguistics.

\bibitem[{Artetxe et~al.(2018)Artetxe, Labaka, Agirre, and
  Cho}]{artetxe2018unsupervised-unmt}
Mikel Artetxe, Gorka Labaka, Eneko Agirre, and Kyunghyun Cho. 2018.
\newblock Unsupervised neural machine translation.
\newblock In \emph{International Conference on Learning Representations}.

\bibitem[{Banerjee and Lavie(2005)}]{banerjee-lavie-2005-meteor}
Satanjeev Banerjee and Alon Lavie. 2005.
\newblock \href {https://www.aclweb.org/anthology/W05-0909} {{METEOR}: An
  automatic metric for {MT} evaluation with improved correlation with human
  judgments}.
\newblock In \emph{Proceedings of the {ACL} Workshop on Intrinsic and Extrinsic
  Evaluation Measures for Machine Translation and/or Summarization}, pages
  65--72, Ann Arbor, Michigan. Association for Computational Linguistics.

\bibitem[{Callison-Burch et~al.(2006)Callison-Burch, Osborne, and
  Koehn}]{callison-burch-etal-2006-evaluating}
Chris Callison-Burch, Miles Osborne, and Philipp Koehn. 2006.
\newblock \href {https://www.aclweb.org/anthology/E06-1032} {Re-evaluating the
  role of {B}leu in machine translation research}.
\newblock In \emph{11th Conference of the {E}uropean Chapter of the Association
  for Computational Linguistics}, Trento, Italy. Association for Computational
  Linguistics.

\bibitem[{Chen et~al.(2019)Chen, Wu, and Zaki}]{chen2020reinforcement}
Yu~Chen, Lingfei Wu, and Mohammed~J Zaki. 2019.
\newblock Reinforcement learning based graph-to-sequence model for natural
  question generation.
\newblock In \emph{International Conference on Learning Representations}.

\bibitem[{Devlin et~al.(2019)Devlin, Chang, Lee, and Toutanova}]{bert}
Jacob Devlin, Ming-Wei Chang, Kenton Lee, and Kristina Toutanova. 2019.
\newblock \href {https://doi.org/10.18653/v1/N19-1423} {{BERT}: Pre-training of
  deep bidirectional transformers for language understanding}.
\newblock In \emph{Proceedings of the 2019 Conference of the North {A}merican
  Chapter of the Association for Computational Linguistics: Human Language
  Technologies, Volume 1 (Long and Short Papers)}, pages 4171--4186,
  Minneapolis, Minnesota. Association for Computational Linguistics.

\bibitem[{Dhole and Manning(2020)}]{dhole-manning-2020-syn-qg}
Kaustubh Dhole and Christopher~D. Manning. 2020.
\newblock \href {https://doi.org/10.18653/v1/2020.acl-main.69} {Syn-{QG}:
  Syntactic and shallow semantic rules for question generation}.
\newblock In \emph{Proceedings of the 58th Annual Meeting of the Association
  for Computational Linguistics}, pages 752--765, Online. Association for
  Computational Linguistics.

\bibitem[{Dong et~al.(2019)Dong, Yang, Wang, Wei, Liu, Wang, Gao, Zhou, and
  Hon}]{dong2019unified-unilm}
Li~Dong, Nan Yang, Wenhui Wang, Furu Wei, Xiaodong Liu, Yu~Wang, Jianfeng Gao,
  Ming Zhou, and Hsiao-Wuen Hon. 2019.
\newblock \href {http://arxiv.org/abs/1905.03197} {Unified language model
  pre-training for natural language understanding and generation}.

\bibitem[{Du et~al.(2017)Du, Shao, and Cardie}]{du-etal-2017-learning}
Xinya Du, Junru Shao, and Claire Cardie. 2017.
\newblock \href {https://doi.org/10.18653/v1/P17-1123} {Learning to ask: Neural
  question generation for reading comprehension}.
\newblock In \emph{Proceedings of the 55th Annual Meeting of the Association
  for Computational Linguistics (Volume 1: Long Papers)}, pages 1342--1352,
  Vancouver, Canada. Association for Computational Linguistics.

\bibitem[{Dzendzik et~al.(2021)Dzendzik, Vogel, and
  Foster}]{dzendzik2021english_qa-dataset}
Daria Dzendzik, Carl Vogel, and Jennifer Foster. 2021.
\newblock English machine reading comprehension datasets: A survey.
\newblock In \emph{Proceedings of the 2021 Conference on Empirical Methods in
  Natural Language Processing (EMNLP)}. Association for Computational
  Linguistics.

\bibitem[{Fabbri et~al.(2020)Fabbri, Ng, Wang, Nallapati, and
  Xiang}]{fabbri-etal-2020-template}
Alexander Fabbri, Patrick Ng, Zhiguo Wang, Ramesh Nallapati, and Bing Xiang.
  2020.
\newblock \href {https://doi.org/10.18653/v1/2020.acl-main.413} {Template-based
  question generation from retrieved sentences for improved unsupervised
  question answering}.
\newblock In \emph{Proceedings of the 58th Annual Meeting of the Association
  for Computational Linguistics}, pages 4508--4513, Online. Association for
  Computational Linguistics.

\bibitem[{Fisch et~al.(2019)Fisch, Talmor, Jia, Seo, Choi, and
  Chen}]{fisch2019mrqa}
Adam Fisch, Alon Talmor, Robin Jia, Minjoon Seo, Eunsol Choi, and Danqi Chen.
  2019.
\newblock {MRQA} 2019 shared task: Evaluating generalization in reading
  comprehension.
\newblock In \emph{Proceedings of 2nd Machine Reading for Reading Comprehension
  (MRQA) Workshop at EMNLP}.

\bibitem[{Gardner et~al.(2017)Gardner, Grus, Neumann, Tafjord, Dasigi, Liu,
  Peters, Schmitz, and Zettlemoyer}]{Gardner2017AllenNLP}
Matt Gardner, Joel Grus, Mark Neumann, Oyvind Tafjord, Pradeep Dasigi,
  Nelson~F. Liu, Matthew Peters, Michael Schmitz, and Luke~S. Zettlemoyer.
  2017.
\newblock \href {http://arxiv.org/abs/arXiv:1803.07640} {Allennlp: A deep
  semantic natural language processing platform}.

\bibitem[{{Graesser} et~al.(2005){Graesser}, {Chipman}, {Haynes}, and
  {Olney}}]{education-2005}
A.~C. {Graesser}, P.~{Chipman}, B.~C. {Haynes}, and A.~{Olney}. 2005.
\newblock \href {https://doi.org/10.1109/TE.2005.856149} {Autotutor: an
  intelligent tutoring system with mixed-initiative dialogue}.
\newblock \emph{IEEE Transactions on Education}, 48(4):612--618.

\bibitem[{Heilman and Smith(2009)}]{Heilman2009Question-noah}
Michael Heilman and Noah~A Smith. 2009.
\newblock Question generation via overgenerating transformations and ranking.
\newblock Technical report, Carnegie-Mellon University.

\bibitem[{Heilman and Smith(2010)}]{heilman-smith-2010-good}
Michael Heilman and Noah~A. Smith. 2010.
\newblock \href {https://www.aclweb.org/anthology/N10-1086} {Good question!
  statistical ranking for question generation}.
\newblock In \emph{Human Language Technologies: The 2010 Annual Conference of
  the North {A}merican Chapter of the Association for Computational
  Linguistics}, pages 609--617, Los Angeles, California. Association for
  Computational Linguistics.

\bibitem[{Hong et~al.(2020)Hong, Kang, Lim, and
  Myaeng}]{kang-etal-2020-handling}
Giwon Hong, Junmo Kang, Doyeon Lim, and Sung-Hyon Myaeng. 2020.
\newblock \href {https://doi.org/10.18653/v1/2020.coling-main.306} {Handling
  anomalies of synthetic questions in unsupervised question answering}.
\newblock In \emph{Proceedings of the 28th International Conference on
  Computational Linguistics}, pages 3441--3448, Barcelona, Spain (Online).
  International Committee on Computational Linguistics.

\bibitem[{Honnibal et~al.(2020)Honnibal, Montani, Van~Landeghem, and
  Boyd}]{spacy}
Matthew Honnibal, Ines Montani, Sofie Van~Landeghem, and Adriane Boyd. 2020.
\newblock \href {https://doi.org/10.5281/zenodo.1212303} {{spaCy:
  Industrial-strength Natural Language Processing in Python}}.

\bibitem[{Joshi et~al.(2017)Joshi, Choi, Weld, and
  Zettlemoyer}]{joshi-etal-2017-triviaqa}
Mandar Joshi, Eunsol Choi, Daniel Weld, and Luke Zettlemoyer. 2017.
\newblock \href {https://doi.org/10.18653/v1/P17-1147} {{T}rivia{QA}: A large
  scale distantly supervised challenge dataset for reading comprehension}.
\newblock In \emph{Proceedings of the 55th Annual Meeting of the Association
  for Computational Linguistics (Volume 1: Long Papers)}, pages 1601--1611,
  Vancouver, Canada. Association for Computational Linguistics.

\bibitem[{Kwiatkowski et~al.(2019)Kwiatkowski, Palomaki, Redfield, Collins,
  Parikh, Alberti, Epstein, Polosukhin, Devlin, Lee, Toutanova, Jones, Kelcey,
  Chang, Dai, Uszkoreit, Le, and
  Petrov}]{kwiatkowski-etal-2019-naturalquestions-nq}
Tom Kwiatkowski, Jennimaria Palomaki, Olivia Redfield, Michael Collins, Ankur
  Parikh, Chris Alberti, Danielle Epstein, Illia Polosukhin, Jacob Devlin,
  Kenton Lee, Kristina Toutanova, Llion Jones, Matthew Kelcey, Ming-Wei Chang,
  Andrew~M. Dai, Jakob Uszkoreit, Quoc Le, and Slav Petrov. 2019.
\newblock \href {https://doi.org/10.1162/tacl_a_00276} {Natural questions: A
  benchmark for question answering research}.
\newblock \emph{Transactions of the Association for Computational Linguistics},
  7:452--466.

\bibitem[{Lewis et~al.(2020)Lewis, Liu, Goyal, Ghazvininejad, Mohamed, Levy,
  Stoyanov, and Zettlemoyer}]{lewis-etal-2020-bart}
Mike Lewis, Yinhan Liu, Naman Goyal, Marjan Ghazvininejad, Abdelrahman Mohamed,
  Omer Levy, Veselin Stoyanov, and Luke Zettlemoyer. 2020.
\newblock \href {https://doi.org/10.18653/v1/2020.acl-main.703} {{BART}:
  Denoising sequence-to-sequence pre-training for natural language generation,
  translation, and comprehension}.
\newblock In \emph{Proceedings of the 58th Annual Meeting of the Association
  for Computational Linguistics}, pages 7871--7880, Online. Association for
  Computational Linguistics.

\bibitem[{Lewis et~al.(2019)Lewis, Denoyer, and
  Riedel}]{lewis-etal-2019-unsupervised-cloze}
Patrick Lewis, Ludovic Denoyer, and Sebastian Riedel. 2019.
\newblock \href {https://doi.org/10.18653/v1/P19-1484} {Unsupervised question
  answering by cloze translation}.
\newblock In \emph{Proceedings of the 57th Annual Meeting of the Association
  for Computational Linguistics}, pages 4896--4910, Florence, Italy.
  Association for Computational Linguistics.

\bibitem[{Li et~al.(2020)Li, Wang, Dong, Wei, and Xu}]{li-etal-2020-harvesting}
Zhongli Li, Wenhui Wang, Li~Dong, Furu Wei, and Ke~Xu. 2020.
\newblock \href {https://doi.org/10.18653/v1/2020.acl-main.600} {Harvesting and
  refining question-answer pairs for unsupervised {QA}}.
\newblock In \emph{Proceedings of the 58th Annual Meeting of the Association
  for Computational Linguistics}, pages 6719--6728, Online. Association for
  Computational Linguistics.

\bibitem[{Lin(2004)}]{lin-2004-rouge}
Chin-Yew Lin. 2004.
\newblock \href {https://www.aclweb.org/anthology/W04-1013} {{ROUGE}: A package
  for automatic evaluation of summaries}.
\newblock In \emph{Text Summarization Branches Out}, pages 74--81, Barcelona,
  Spain. Association for Computational Linguistics.

\bibitem[{Loshchilov and Hutter(2019)}]{adamw}
Ilya Loshchilov and Frank Hutter. 2019.
\newblock \href {https://openreview.net/forum?id=Bkg6RiCqY7} {Decoupled weight
  decay regularization}.
\newblock In \emph{International Conference on Learning Representations}.

\bibitem[{Ma et~al.(2020)Ma, Zhu, Zhou, and Li}]{ma2020improving}
Xiyao Ma, Qile Zhu, Yanlin Zhou, and Xiaolin Li. 2020.
\newblock Improving question generation with sentence-level semantic matching
  and answer position inferring.
\newblock In \emph{Proceedings of the AAAI Conference on Artificial
  Intelligence}, volume~34, pages 8464--8471.

\bibitem[{Narayan et~al.(2018)Narayan, Cohen, and
  Lapata}]{narayan-etal-2018-xsum}
Shashi Narayan, Shay~B. Cohen, and Mirella Lapata. 2018.
\newblock \href {https://doi.org/10.18653/v1/D18-1206} {Don{'}t give me the
  details, just the summary! topic-aware convolutional neural networks for
  extreme summarization}.
\newblock In \emph{Proceedings of the 2018 Conference on Empirical Methods in
  Natural Language Processing}, pages 1797--1807, Brussels, Belgium.
  Association for Computational Linguistics.

\bibitem[{Narayan et~al.(2020)Narayan, Simoes, Ma, Craighead, and
  Mcdonald}]{narayan2020qurious-maji}
Shashi Narayan, Gonçalo Simoes, Ji~Ma, Hannah Craighead, and Ryan Mcdonald.
  2020.
\newblock \href {http://arxiv.org/abs/2004.11026} {Qurious: Question generation
  pretraining for text generation}.

\bibitem[{Papineni et~al.(2002)Papineni, Roukos, Ward, and
  Zhu}]{papineni-etal-2002-bleu}
Kishore Papineni, Salim Roukos, Todd Ward, and Wei-Jing Zhu. 2002.
\newblock \href {https://doi.org/10.3115/1073083.1073135} {{B}leu: a method for
  automatic evaluation of machine translation}.
\newblock In \emph{Proceedings of the 40th Annual Meeting of the Association
  for Computational Linguistics}, pages 311--318, Philadelphia, Pennsylvania,
  USA. Association for Computational Linguistics.

\bibitem[{Puri et~al.(2020)Puri, Spring, Shoeybi, Patwary, and
  Catanzaro}]{puri-etal-2020-training-synthetic}
Raul Puri, Ryan Spring, Mohammad Shoeybi, Mostofa Patwary, and Bryan Catanzaro.
  2020.
\newblock \href {https://doi.org/10.18653/v1/2020.emnlp-main.468} {Training
  question answering models from synthetic data}.
\newblock In \emph{Proceedings of the 2020 Conference on Empirical Methods in
  Natural Language Processing (EMNLP)}, pages 5811--5826, Online. Association
  for Computational Linguistics.

\bibitem[{Rajpurkar et~al.(2016)Rajpurkar, Zhang, Lopyrev, and
  Liang}]{rajpurkar-etal-2016-squad1.1}
Pranav Rajpurkar, Jian Zhang, Konstantin Lopyrev, and Percy Liang. 2016.
\newblock \href {https://doi.org/10.18653/v1/D16-1264} {{SQ}u{AD}: 100,000+
  questions for machine comprehension of text}.
\newblock In \emph{Proceedings of the 2016 Conference on Empirical Methods in
  Natural Language Processing}, pages 2383--2392, Austin, Texas. Association
  for Computational Linguistics.

\bibitem[{Reiter(2018)}]{10.1162/coli_a_00322}
Ehud Reiter. 2018.
\newblock \href {https://doi.org/10.1162/coli_a_00322} {{A Structured Review of
  the Validity of BLEU}}.
\newblock \emph{Computational Linguistics}, 44(3):393--401.

\bibitem[{Saha et~al.(2018)Saha, Aralikatte, Khapra, and
  Sankaranarayanan}]{saha-etal-2018-duorc}
Amrita Saha, Rahul Aralikatte, Mitesh~M. Khapra, and Karthik Sankaranarayanan.
  2018.
\newblock \href {https://doi.org/10.18653/v1/P18-1156} {{D}uo{RC}: Towards
  complex language understanding with paraphrased reading comprehension}.
\newblock In \emph{Proceedings of the 56th Annual Meeting of the Association
  for Computational Linguistics (Volume 1: Long Papers)}, pages 1683--1693,
  Melbourne, Australia. Association for Computational Linguistics.

\bibitem[{Seo et~al.(2016)Seo, Kembhavi, Farhadi, and
  Hajishirzi}]{seo2016bidirectional_bidaf}
Minjoon Seo, Aniruddha Kembhavi, Ali Farhadi, and Hannaneh Hajishirzi. 2016.
\newblock Bidirectional attention flow for machine comprehension.
\newblock \emph{arXiv preprint arXiv:1611.01603}.

\bibitem[{Shakeri et~al.(2020)Shakeri, Nogueira~dos Santos, Zhu, Ng, Nan, Wang,
  Nallapati, and Xiang}]{shakeri-etal-2020-end-wangzhiguo}
Siamak Shakeri, Cicero Nogueira~dos Santos, Henghui Zhu, Patrick Ng, Feng Nan,
  Zhiguo Wang, Ramesh Nallapati, and Bing Xiang. 2020.
\newblock \href {https://doi.org/10.18653/v1/2020.emnlp-main.439} {End-to-end
  synthetic data generation for domain adaptation of question answering
  systems}.
\newblock In \emph{Proceedings of the 2020 Conference on Empirical Methods in
  Natural Language Processing (EMNLP)}, pages 5445--5460, Online. Association
  for Computational Linguistics.

\bibitem[{Song et~al.(2019)Song, Tan, Qin, Lu, and Liu}]{song2019mass}
Kaitao Song, Xu~Tan, Tao Qin, Jianfeng Lu, and Tie-Yan Liu. 2019.
\newblock Mass: Masked sequence to sequence pre-training for language
  generation.
\newblock In \emph{International Conference on Machine Learning}, pages
  5926--5936. PMLR.

\bibitem[{Sun et~al.(2018)Sun, Liu, Lyu, He, Ma, and
  Wang}]{sun-etal-2018-answer-yanjunma}
Xingwu Sun, Jing Liu, Yajuan Lyu, Wei He, Yanjun Ma, and Shi Wang. 2018.
\newblock \href {https://doi.org/10.18653/v1/D18-1427} {Answer-focused and
  position-aware neural question generation}.
\newblock In \emph{Proceedings of the 2018 Conference on Empirical Methods in
  Natural Language Processing}, pages 3930--3939, Brussels, Belgium.
  Association for Computational Linguistics.

\bibitem[{Sutskever et~al.(2014)Sutskever, Vinyals, and Le}]{NIPS2014_seq2seq}
Ilya Sutskever, Oriol Vinyals, and Quoc~V Le. 2014.
\newblock \href
  {https://proceedings.neurips.cc/paper/2014/file/a14ac55a4f27472c5d894ec1c3c743d2-Paper.pdf}
  {Sequence to sequence learning with neural networks}.
\newblock In \emph{Advances in Neural Information Processing Systems},
  volume~27. Curran Associates, Inc.

\bibitem[{Trischler et~al.(2017)Trischler, Wang, Yuan, Harris, Sordoni,
  Bachman, and Suleman}]{trischler-etal-2017-newsqa}
Adam Trischler, Tong Wang, Xingdi Yuan, Justin Harris, Alessandro Sordoni,
  Philip Bachman, and Kaheer Suleman. 2017.
\newblock \href {https://doi.org/10.18653/v1/W17-2623} {{N}ews{QA}: A machine
  comprehension dataset}.
\newblock In \emph{Proceedings of the 2nd Workshop on Representation Learning
  for {NLP}}, pages 191--200, Vancouver, Canada. Association for Computational
  Linguistics.

\bibitem[{Tsatsaronis et~al.(2015)Tsatsaronis, Balikas, Malakasiotis, Partalas,
  Zschunke, Alvers, Weissenborn, Krithara, Petridis, Polychronopoulos
  et~al.}]{bio_asq}
George Tsatsaronis, Georgios Balikas, Prodromos Malakasiotis, Ioannis Partalas,
  Matthias Zschunke, Michael~R Alvers, Dirk Weissenborn, Anastasia Krithara,
  Sergios Petridis, Dimitris Polychronopoulos, et~al. 2015.
\newblock An overview of the bioasq large-scale biomedical semantic indexing
  and question answering competition.
\newblock \emph{BMC bioinformatics}, 16(1):1--28.

\bibitem[{Wang and Jiang()}]{wang2016machine_match-lstm}
Shuohang Wang and Jing Jiang.
\newblock Machine comprehension using match-lstm and answer pointer.(2017).
\newblock In \emph{ICLR 2017: International Conference on Learning
  Representations, Toulon, France, April 24-26: Proceedings}, pages 1--15.

\bibitem[{Wolf et~al.(2019)Wolf, Debut, Sanh, Chaumond, Delangue, Moi, Cistac,
  Rault, Louf, Funtowicz, and Brew}]{Wolf2019HuggingFacesTS}
Thomas Wolf, Lysandre Debut, Victor Sanh, Julien Chaumond, Clement Delangue,
  Anthony Moi, Pierric Cistac, Tim Rault, R'emi Louf, Morgan Funtowicz, and
  Jamie Brew. 2019.
\newblock Huggingface's transformers: State-of-the-art natural language
  processing.
\newblock \emph{ArXiv}, abs/1910.03771.

\bibitem[{Xie et~al.(2020)Xie, Pan, Wang, Kan, and
  Feng}]{xie-etal-2020-exploring}
Yuxi Xie, Liangming Pan, Dongzhe Wang, Min-Yen Kan, and Yansong Feng. 2020.
\newblock \href {https://doi.org/10.18653/v1/2020.coling-main.228} {Exploring
  question-specific rewards for generating deep questions}.
\newblock In \emph{Proceedings of the 28th International Conference on
  Computational Linguistics}, pages 2534--2546, Barcelona, Spain (Online).
  International Committee on Computational Linguistics.

\bibitem[{Zhang and Bansal(2019)}]{zhang-bansal-2019-addressing}
Shiyue Zhang and Mohit Bansal. 2019.
\newblock \href {https://doi.org/10.18653/v1/D19-1253} {Addressing semantic
  drift in question generation for semi-supervised question answering}.
\newblock In \emph{Proceedings of the 2019 Conference on Empirical Methods in
  Natural Language Processing and the 9th International Joint Conference on
  Natural Language Processing (EMNLP-IJCNLP)}, pages 2495--2509, Hong Kong,
  China. Association for Computational Linguistics.

\end{thebibliography}
\bibliographystyle{acl_natbib}

\newpage

\appendix

\section{Appendix}
\label{sec:appendix}

\subsection{Effects of Different Beam Size}

\begin{figure}[h]
    \centering
    \includegraphics[scale=0.46]{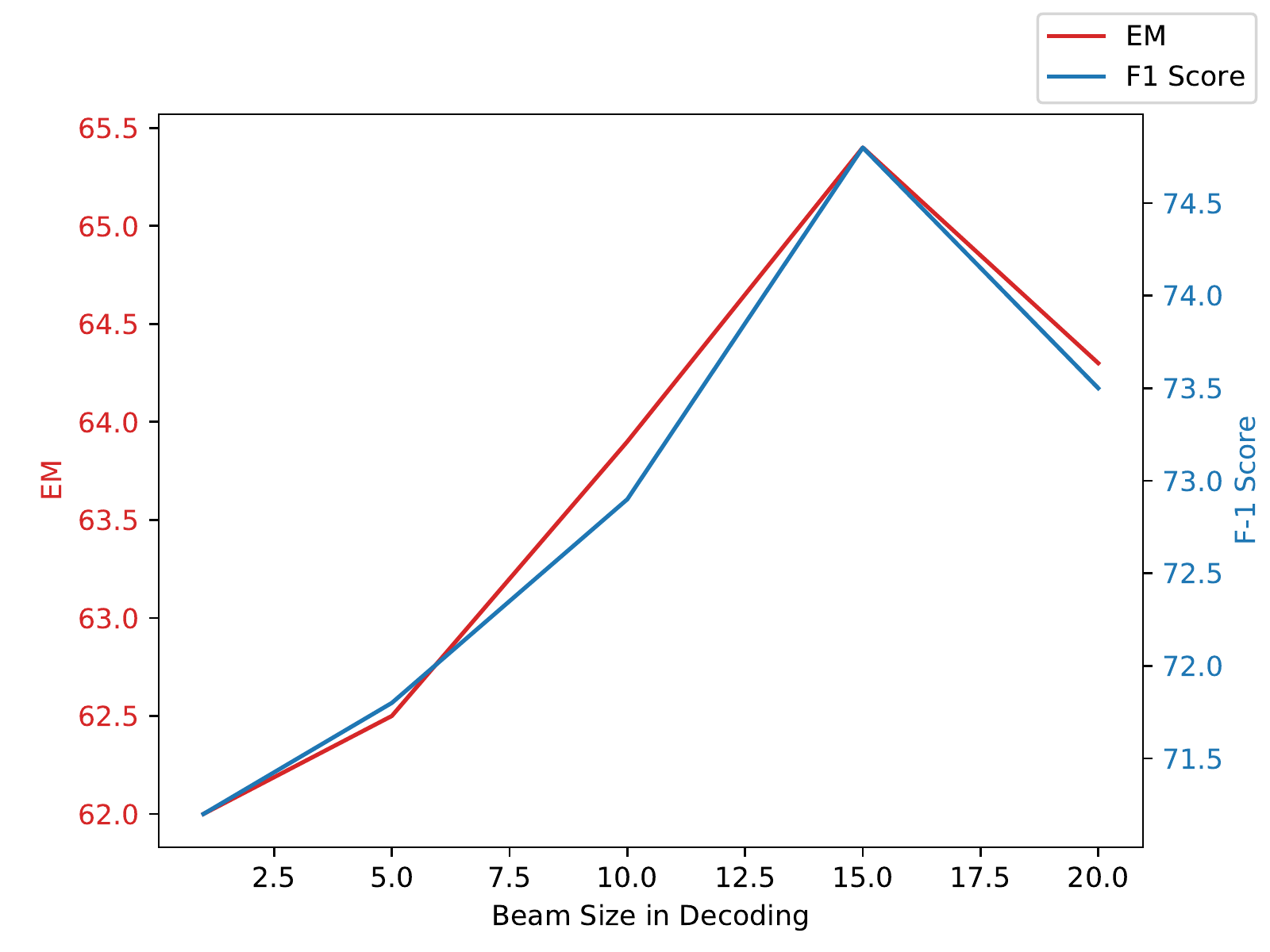}
    \caption{Experimental results of the effects of using different beam-size in decoding process when generating synthetic questions.}
    \label{fig:effect_of_beam_size}
\end{figure}

We also study the effects of different beam size in generating synthetic questions to the performance of downstream QA task. Experiments are conducted on SQuAD1.1 dev set using BERT-large, questions in the synthetic QA data are generated with different beam size using the same BART-QG model. The experimental results in Figure~\ref{fig:effect_of_beam_size} show that the beam size is an important factor affecting the performance of unsupervised QA, the largest margin between the highest score~(beam-15) and the lowest score~(beam-1) in 
Figure~\ref{fig:effect_of_beam_size} is close to 4 points on EM and F-1 score.

\subsection{Question Type Distribution}

\begin{figure}[h]
    \centering
    \includegraphics[scale=0.46]{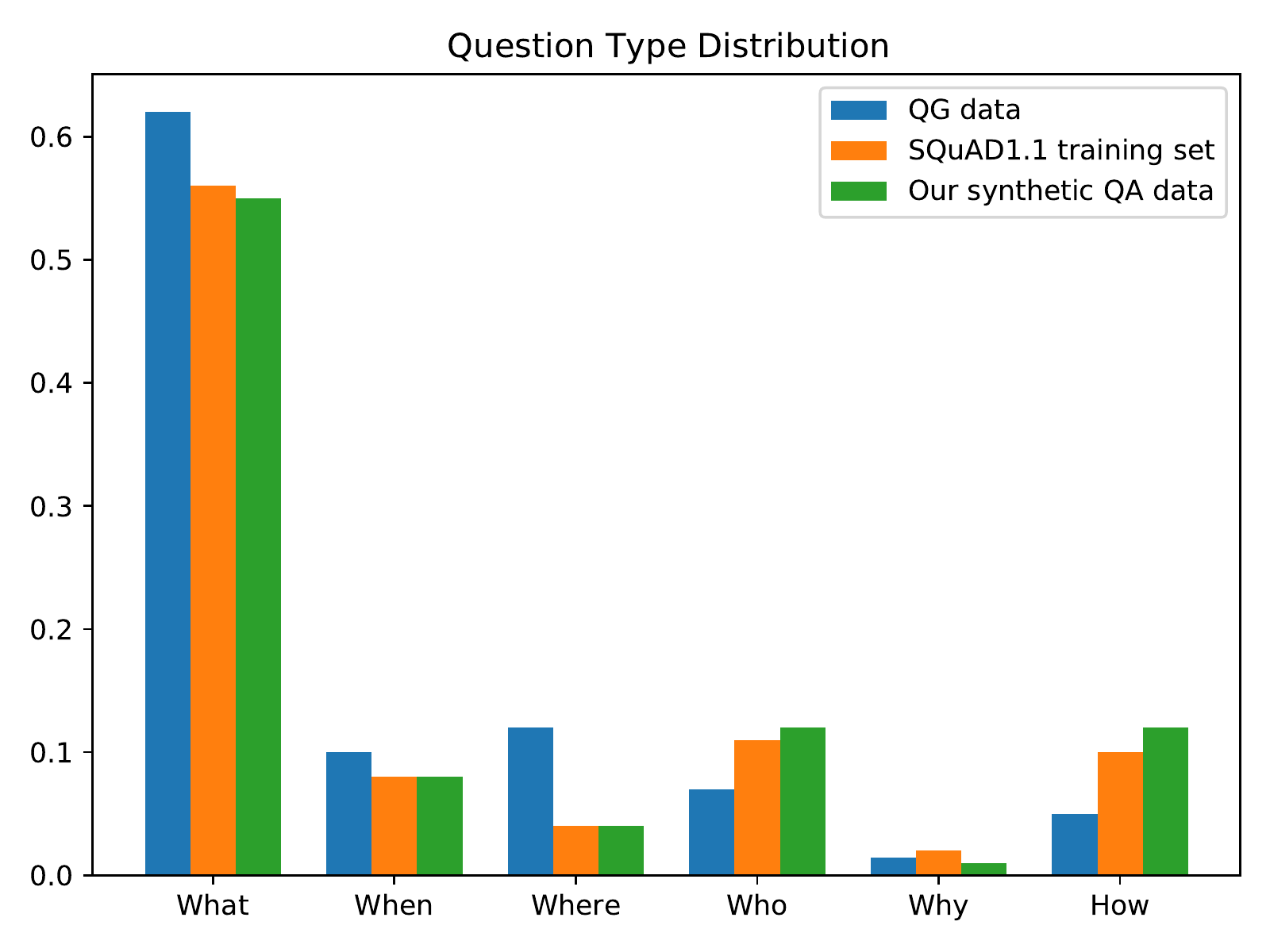}
    \caption{Question type distribution}
    \label{fig:question_type_distribution}
\end{figure}

We show the distribution of question types of QG data described in Section~4.1.1, training set of SQuAD1.1 and our synthetic QA data in Section~4.1.2 in Figure~\ref{fig:question_type_distribution}, question types are defined as \textit{What, When, Where, Who, Why, How}. The QG data has more \textit{what, when, where} questions, indicating the existence of more SRL arguments associated with such question types in the summary sentences.

\subsection{Generated QA Examples}
Some Wikipedia-based \textit{<passage, answer, question>} examples generated by our BART-QG model are shown in Table~\ref{qg_results_0}, Table~\ref{qg_results_1} and Table~\ref{qg_results_2}.
\begin{table*}
    \centering

  \begin{tabular}{|p{6cm}|p{4cm}|p{5cm}|}

    \hline
    
    Passage & Answer & Question \\
    
    \hline
    
At a professional level, most matches produce only a few goals.  For example, the 2005–06 season of the English Premier League produced an average of 2.48 goals per match.  The Laws of the Game do not specify any player positions other than goalkeeper, but a number of specialised roles have evolved.

    & the 2005–06 season & when did the english football team produce an average of 2.49 goals per match  , according to the laws of the game ? \\

    \hline
    The Hebrew Book Week is held each June and features book fairs, public readings, and appearances by Israeli authors around the country.  During the week, Israel's top literary award, the Sapir Prize, is presented.
    
    & The Hebrew Book Week & what is held every june to celebrate the publication of books in hebrew ? \\
    
    \hline
    
    On December 12, 2016, Senate Majority Leader Republican Mitch McConnell expressed confidence in U.S. intelligence.  McConnell added that investigation of Russia's actions should be bipartisan and held by the Senate Intelligence Committee.  The next day, Senate Intelligence Committee Chairman Richard Burr (R-NC) and Vice Chairman Mark Warner (D-VA) announced the scope of the committee's . 
    
    & Republican Mitch McConnell & which republican  has called for a special committee to investigate russia 's alleged meddling in the 2016 presidential election ? \\
    
    \hline
    Meanwhile, the Soho Mint struck coins for the East India Company, Sierra Leone and Russia, while producing high-quality planchets, or blank coins, to be struck by national mints elsewhere.  The firm sent over 20 million blanks to Philadelphia, to be struck into cents and half-cents by the United States Mint —Mint Director Elias Boudinot found them to be "perfect and beautifully polished".
    & Elias Boudinot & who has been working for a company that made coins for the us mint ? \\
    
    \hline
    
    \end{tabular}%
    \caption{Some generated QA examples.
      \label{qg_results_0}%
    }

\end{table*}

\begin{table*}
    \centering

  \begin{tabular}{|p{6cm}|p{4cm}|p{5cm}|}

    \hline
    
    Passage & Answer & Question \\
    
    \hline
    
    In March 2008 as part of the annual budget, the government introduced several laws to amend the Immigration and Refugee Protection Act.  The changes would have helped to streamline immigrant application back-up, to speed up application for skilled workers and to rapidly reject other ones that are judged not admissible by immigration officers.  Immigrant applications had risen to a high of 500,000, creating a delay of up to six months for an application to be processed. 
    
    & March 2008 & when did the uk introduce new immigration laws ? \\
    
    \hline
    
    The other group members as far back as 1996 had noticed Paddy Clancy's unusual mood swings.  In the spring of 1998 the cause was finally detected; Paddy had a brain tumor as well as lung cancer.  His wife waited to tell him about the lung cancer, so as not to discourage him when he had a brain operation.  
    
    & the spring of 1998 & in what time was paddy diagnosed with lung cancer ? \\
    
    \hline
      In 1365 officials were created to supervise the fish market in the town, whilst illegal fishing and oyster cultivation was targeted by the bailiffs in an edict from 1382, which prohibited the forestalling of fish by blocking the river, the dredging of oysters out of season and the obstructing of the river.  Colchester artisans included clockmakers, who maintained clocks in church towers across north Essex and Suffolk.

    & north Essex & where were hundreds of clocks made by local artisans ? \\
    
    \hline
    Badge numbers for Sheriffs and Deputies consist of a prefix number, which represents the county number, followed by a one to three digit number, which represents the Sheriff's or Deputy's number within that specific office.  The Sheriff's badge number in each county is always \#1.  So the Sheriff from Bremer County would have an ID number of 9-1 (9 is the county number for Bremer County and 1 is the number for the Sheriff).

    & The Sheriff's badge number & what is the number used to identify the sheriff in each county ? \\
    
    \hline
    
    \end{tabular}%
    \caption{Some generated QA examples.
      \label{qg_results_1}%
    }

\end{table*}

\begin{table*}
    \centering

  \begin{tabular}{|p{6cm}|p{4cm}|p{5cm}|}

    \hline
    
    Passage & Answer & Question \\
    
    \hline
    
Appian wrote that Calpurnius Piso was sent as a commander to Hispania because there were revolts.  The following year Servius Galba was sent without soldiers because the Romans were busy with Cimbrian War and a slave rebellion in Sicily (the [Third Servile War], 104-100 BC).  In the former war the Germanic tribes of the Cimbri and the Teutones migrated around Europe and invaded territories of allies of Rome, particularly in southern France, and routed the Romans in several battles until their final defeat.

    & Calpurnius Piso & who was sent to the south of italy to fight for the roman empire ? \\
    
    \hline
    
The parish churches of Sempringham, Birthorpe, Billingborough, and Kirkby were already appropriated.  Yet in 1247, Pope Innocent IV granted to the master the right to appropriate the church of Horbling, because there were 200 women in the priory who often lacked the necessaries of life.  The legal expenses of the order at the papal curia perhaps accounted for their poverty.

    & 200 & there were how many women in the priory of horbling in the 12th century ? \\
    
    \hline
"Jerry West is the reason I came to the Lakers", O'Neal later said.  They used their 24th pick in the draft to select Derek Fisher.  During the 1996–97 season, the team traded Cedric Ceballos to Phoenix for Robert Horry.  O'Neal led the team to a 56–26 record, their best effort since 1990–91, despite missing 31 games due to a knee injury.  O'Neal averaged 26.2 ppg and 12.5 rpg and finished third in the league in blocked shots (2.88 bpg) in 51 games.
    
    & the 1996–97 season & when do the phoenix suns begin  with a trade to the los angeles clippers ? \\
    
    \hline
Finnish popular music also includes various kinds of dance music; tango, a style of Argentine music, is also popular.  One of the most productive composers of popular music was Toivo Kärki, and the most famous singer Olavi Virta (1915–1972).  Among the lyricists, Sauvo Puhtila (1928–2014), Reino Helismaa (died 1965) and Veikko "Vexi" Salmi are a few of the most notable writers.  The composer and bandleader Jimi Tenor is well known for his brand of retro-funk music.

    & Reino Helismaa & who has been hailed as one of finland 's most important writers ? \\
    
    \hline
    
    \end{tabular}%
    \caption{Some generated QA examples.
      \label{qg_results_2}%
    }

\end{table*}

\end{document}